\documentclass[pdflatex,sn-mathphys]{sn-jnl}

\jyear{2021}%
\usepackage{makecell}
\usepackage{caption}
\usepackage{amsmath}
\usepackage{float}

\theoremstyle{thmstyleone}%
\theoremstyle{thmstyletwo}%

\theoremstyle{thmstylethree}%

\begin{document}

\title[Article Title]{Detecting Anomalies using Generative Adversarial Networks on Images}

\author*[1]{\fnm{Rushikesh} \sur{Zawar}}\email{f20170977@pilani.bits-pilani.ac.in
}

\author[1]{\fnm{Krupa} \sur{Bhayani}}\email{f20180844@pilani.bits-pilani.ac.in}

\author[2]{\fnm{Neelanjan} \sur{Bhowmik}}\email{neelanjan.bhowmik@durham.ac.uk}
\author[1]{\fnm{Kamlesh} \sur{Tiwari}}\email{kamlesh.tiwari@pilani.bits-pilani.ac.in}
\author[3]{\fnm{Dhiraj}}\email{dhiraj@ceeri.res.in}

\affil*[1]{\orgname{Birla Institute of Technology and Science, Pilani}, \orgaddress{\city{Pilani}, \country{India}}}

\affil*[2]{\orgname{Durham University, United Kingdom.}}

\affil*[3]{\orgname{Central Electronics \& Engineering
Research Institute}, \orgaddress{ \country{India}}}

\abstract{Automatic detection of anomalies such as weapons or threat objects in baggage security, or detecting impaired items in industrial production is an important computer vision task demanding high efficiency and accuracy. Most of the available data in the anomaly detection task is imbalanced as the number of positive/anomalous instances is sparse. Inadequate availability of the data makes training of a deep neural network architecture for anomaly detection challenging. This paper proposes a novel Generative Adversarial Network (GAN) based model for anomaly detection. It uses normal (non-anomalous) images to learn about the normality based on which it detects if an input image contains an anomalous/threat object. The proposed model uses a generator with an encoder-decoder network having dense convolutional skip connections for enhanced reconstruction and to capture the data distribution. A self-attention augmented discriminator is used having the ability to check the consistency of detailed features even in distant portions. We use spectral normalisation to facilitate stable and improved training of the GAN. Experiments are performed on three datasets, viz. CIFAR-10, MVTec AD (for industrial applications) and SIXray (for X-ray baggage security). On the MVTec AD and SIXray datasets, our model achieves an improvement of upto 21\% and 4.6\%, respectively, in Area Under Curve (AUC) of the Receiver Operating Characteristic (ROC) curve, outperforming prior works. Due to the sensitivity to the false negatives of the X-ray baggage screening task, additionally, we obtain upto 11\% improvement in Recall on the SIXray dataset compared to the state-of-the-art anomaly detection models using GANs.}

\maketitle
\vspace*{-0.4cm}
\section{Introduction}
\begin{figure}[!htb]
\centering
\includegraphics[width=\linewidth]{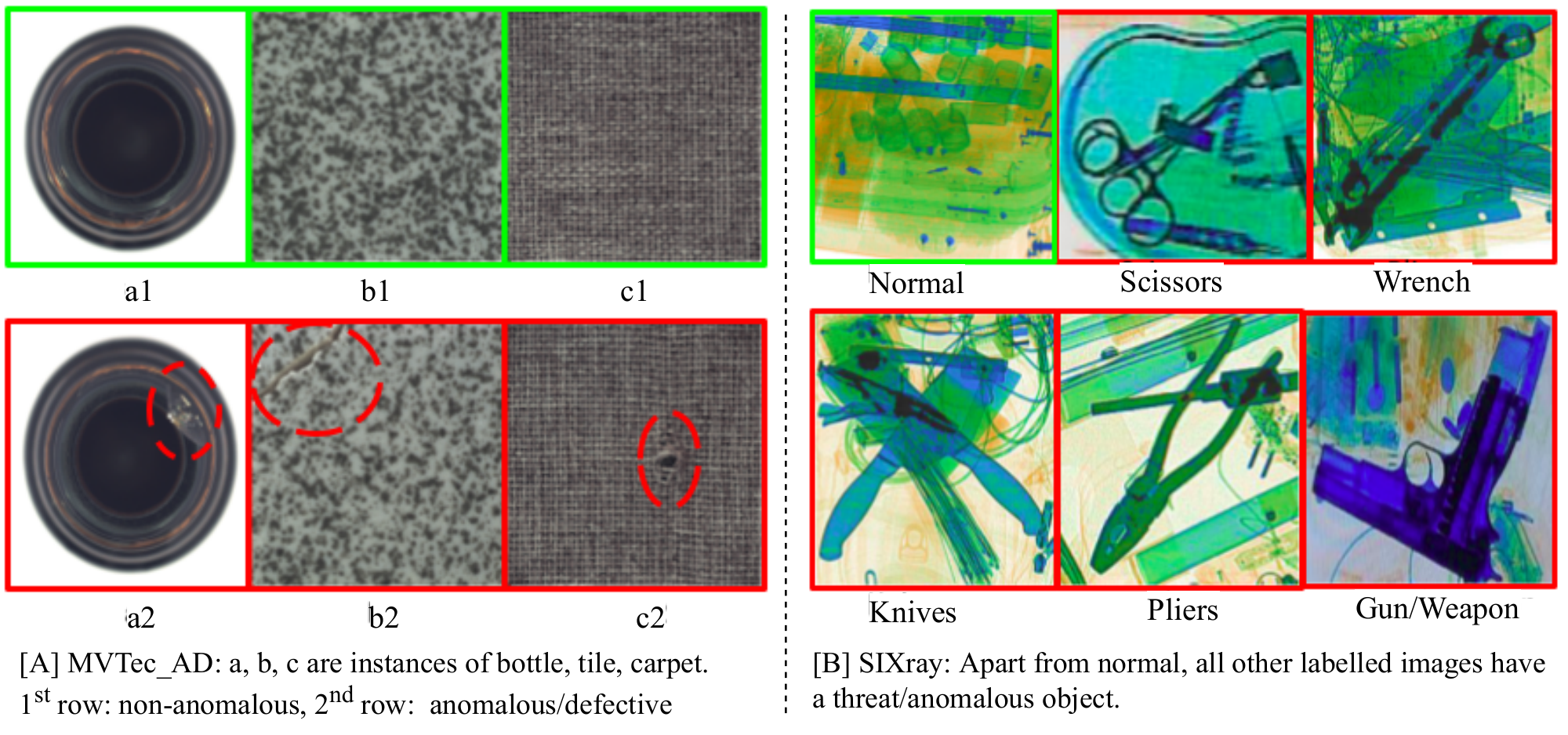} 
\vspace*{-0.4cm}
\caption{Exampler images from MVTec\_AD [A] and SIXray [B] dataset.}
\label{fig:datasets}
\vspace*{-0.4cm}
\end{figure}

\vspace*{-0.2cm}
\label{sec:intro}
   Anomaly detection is an important topic in industries \cite{kammerer2019anomaly}, security \cite{griffin2018unexpected} , medical diagnosis \cite{schlegl2017unsupervised}, $etc.$, where there is a need to increase automation for better efficiency. Image analysis using Artificial Intelligence (AI) is one of the non-destructive way to recognize a potential threat. The data in these fields can be high dimensional and biased. This creates an imbalance in the dataset, where data or images of some classes might not be enough to train deep convolutional neural network architectures. With recent advancements where real data is readily being acquired, the lack or lesser availability of anomalous instances still persists. In baggage security, items/objects such as guns, pliers, knives, scissors, $etc$. are considered anomaly. In industries, a defective product is considered an anomaly to filter out the perfect products. Figure \ref{fig:datasets}, shows the instances of anomalous and non-anomalous images in both industrial production (MVTec\_AD) dataset and both X-ray baggage security (SIXray)  respectively.

    \noindent Our work aims to build a model that can identify if a piece of baggage passed through a security check has a threat or anomalous object. This can be used to make smart systems to
    reduce the manpower required for various tasks that can be automatized.  This automation
    will reduce the laborious work of humans to only special cases where human expertise is
    needed.  In the case of baggage security, a system will call for human help only for suspicious bags. Similarly, our model can be used for detecting defective products in industries.

    \noindent Over the years, to detect anomalies in images various machine learning and deep learning techniques have been used. In classification, methods like SVM \cite{erfani2016high}, K-nearest neighbors \cite{altman1992introduction}, with some feature extractors like Scale Invariant Feature Transform (SIFT) \cite{lowe2004distinctive}, Speeded Up Robust Features (SURF) \cite{bay2006surf}, $etc.$ have been used on x-ray images. Many of these methods have been combined with various object detection and segmentation methods. 
    However, using standard learning algorithms doesn't solve all problems due to imbalance present in the classes of datasets, as they cannot train supervised learning models well enough. To overcome this, recently unsupervised methods like Generative Adversarial Networks(GAN) are being used \cite{akcay2018ganomaly, akcay2020towards}. GAN architecture was first proposed in 2014 \cite{goodfellow2014generative}. The two major parts of this network are the Generator and Discriminator, which are trained in a  zero-sum game, where loss and gain of each are balanced by the loss-gain of other and eventually we let them come at an equilibrium. The Generator is to capture the distribution of the input dataset for a given class label, by predicting features or images from a hidden representation.The Discriminator classifies as real or fake, based on the given features or images.
    \vspace{0.1cm}

    \noindent \section{Related Works:} Anomaly detection is used in various domains like video analysis, medical image analysis, industrial production, remote sensing, $etc$. In the paper DCGAN \cite{radford2015unsupervised} its shown that deep convolutional GANs are capable of capturing semantic image content. In \cite{7280790} a dictionary of filters is used to detect anomalous regions using convolutional sparse models. \cite{schlegl2019f}. Using GANs for anomaly detection is a popular approach in the field of research after Ano-GAN \cite{schlegl2017unsupervised}  developed this concept, where the GAN is trained only on the non-anomalous instances. This method was based on the hypothesis that the distribution of the data can be represented by the latent vector generated by the GAN. For this a Generator and discriminator are trained but using only normal images. This way the latent vector would represent the distribution of the normal data. After these networks are trained, their weights are freezed and are used on a given image and remap the latent vector. The output of the discriminator gives a probability that the given image came from real data  than from the Generator. This is used in the inference to calculate the anomaly score based on which an image is classified as anomalous or non-anonmalous. Residual scores for test images as well as fake images generated by GAN are generated and compared. The Network will detect anomalies by defining thresholds for these scores of both classes. This training process has inspired many recent methods. However, Ano-GAN, required some optimization steps for every new input that lead to overall bad test-time performance. BiGAN \cite{donahue2016adversarial} has a generator that maps latent samples to generated data and there's also inverse mapping from data to latent representation.
    Based on BiGAN, EGBAD \cite{zenati2018efficient} improved the speed of the model by avoiding the computationally expensive step of recovering a latent representation at test time. GANomaly \cite{akcay2018ganomaly} introduced the combination of GAN and auto-encoder and was found to be very effective. The computations are reduced as anomaly scores of test images can be directly compared with fake images without iterations. In 2019, Samet \cite{akccay2019skip} proposed an improved model of GANomaly, $viz.$ Skip-ganomaly. The Skip-ganomaly architecture where skip-connections are added to the generator network, increasing image reconstruction ability. The performance of Skip-ganomaly is more stable than that of AnoGAN and GANomaly. But still, there are some cases where GANs fail to perform well which might be due to modal collapse during the training process. Our proposed method tries to tackle this issue as well as improve the overall performance of the network.
    
    \vspace{0.2cm}
    \noindent The main contributions of this paper are as follows:

    \noindent 1. \emph{Anomaly detection}: A novel model based on GANs, where the Generator(G) has better reconstruction capability with its dense convolutional skip connections. The attention augmented Discriminator(D), is capable of checking consistency even in distant features. Both G \& D, are stably trained using spectral normalization combating mode collapse.

    \noindent 2. \emph{Efficacy}: Our model that has better performance against prior state-of-the-art approaches in CIFAR-10, SIXray and MVTec\_AD datasets.\cite{akccay2019skip, akcay2018ganomaly, schlegl2017unsupervised, tang2020anomaly}. All these datasets have different types of images, especially SIXray which is a Xray dataset. The model is independent of type of images and thus can be adapted to different datasets easily.
    
    
    \noindent 3. \emph{Lesser False Negatives}: Our work considers and also gives the top performance in reducing false negatives in SIXray dataset, as it's an important factor in sensitive field of baggage security applications. These fields are sensitive to false negatives as letting a threat object or damaged product go undetected can be much more costly as compared to letting a normal bag or a good product get detected as anomalous or faulty.\\
    
    \noindent Our work focuses and performs experiments on different datasets which shows that its just not restricted to X-ray baggage or industrial applications, but rather is useful in both of them. Also there is no dataset or image type specific preprocessing, which we believe makes this quite generalizable and thus can be easily used in other domains also. The improvement in the reconstruction of the image by the GAN as comapred to other methods which don't use dense skip connections was quite apparent from the reconstructed images (images shown in supplementary material). The consistency in the AUC of our model as seen in table \ref{recall} across the different dataset sets, shows that our model performs quite well even with smaller size of dataset, which reduces its training requirements. Even within each set, different random combinations were used to avoid overfitting.

    \noindent The detailed architecture of our model is explained in section \ref{sec:propsed}

\vspace*{-0.2cm}
    \section{Proposed Approach}
    \label{sec:propsed}

    \vspace*{-0.1cm}
    
    The lack of a balanced dataset in Anomaly Detection field requires a solution that can mitigate this problem. Generative methods are one of the top methods in such cases. This led us to use GANs. Using GANomaly \cite{akcay2018ganomaly} as the based model, as it is one of the simplest GAN model, different problems were observed which are explained ahead.This model helped us get insights into where simple GAN models are lacking. Solving the reconstruction problem, the diminishing and exploding gradient and the only local level relations in the images led to the different additions and modifications in the architecture. These observations were across datasets and not specific to any particular dataset.  Our method is an unary classification method that uses GAN and we train the model on non-anomalous images $i.e$. images without any anomalous object or any abnormality. However, after training, we test on both the non-anomalous and anomalous images. 
    The training objective of our model is to capture the distribution of the training dataset in both image and hidden latent vector space, as the latent representation is a unique representation of the image it has been built from. This helps the model learn the features of non-anomalous images that are unique to them. We used a measure called Anomaly Score, to rate an input if it is anomalous or not, where the value is directly proportional to the likelihood of it having an anomaly. 
    \\
  
    \begin{figure*}
        \centering
        \includegraphics[width=4.5in, height=2.0in]{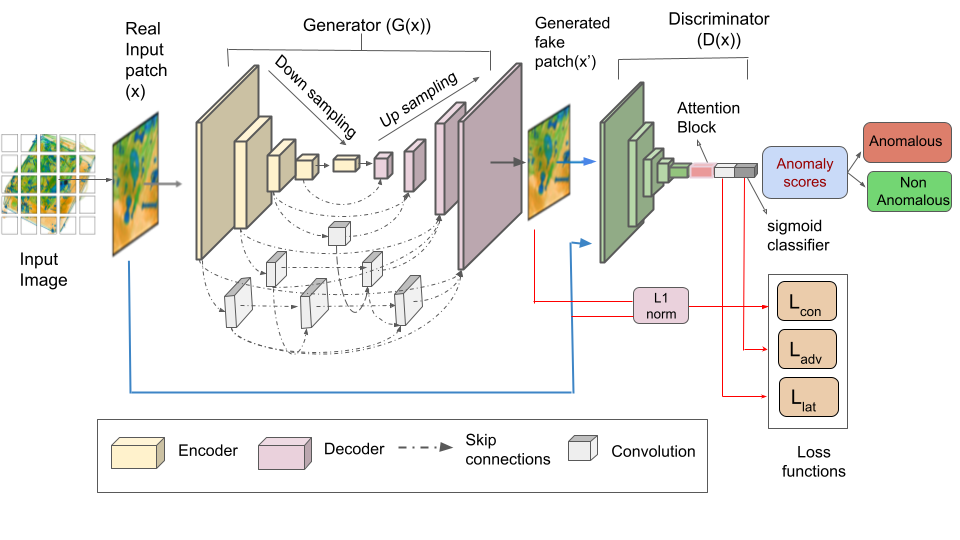}
        \vspace*{-0.5cm}
        \caption{\small Overview of proposed anomaly detection model}
        \label{fig:overv}
        \vspace*{-0.2cm}
    \end{figure*}
    
    \vspace*{-0.2cm}
    \noindent As shown in Figure \ref{fig:overv}, the model has the first step extracting patches from images in a grid-wise fashion. These patches are then fed to the model. The generator(G) comes first followed by the discriminator(D). The G generates images, and both the original and generated image are fed to the D, which learns to distinguish the original from the generated one. During testing, a similar flow is followed. The image is fed to the model and an image is generated. Both the original input image and generated image are given to the D. The data obtained from this process is then used to calculate the anomaly score, as explained in section \ref{sec:propsed}. Based on this anomaly score the image is classified as non-anomalous or anomalous. The architecture and functions of both G and D are described below. The use of spectral-normalization for stabilizing the training of GANs and also the use of attention to increase the capability of the discriminator to check the consistency of detailed features in distant portions are explained in the following segments.

    \noindent \textbf{Generator Network(G):} The network begins with an encoder sub-network followed by a decoder sub-network.  Motivated by \cite{zhou2018unet++, huang2017densely}, the Encoder and Decoder are also connected by a dense convolutional skip connection block. These dense convolutional blocks have different sizes, depending on the layers that they are connecting. In the dense blocks at a given level, the output from the previous convolutional layer of the same dense block is concatenated with the output from the which is upsampled from the convolutional layer below it. The aim behind this connection is to decrease the semantic gap between the feature maps from the encoder and decoder, which would help and bring about better reconstruction as both local and global information is preserved through these skip connections. The encoder maps the input image to a low dimensional latent representation by downsampling and tries to capture its distribution. Contrary to the encoder, the decoder upsamples the latent representation to reconstruct the image. The block in both encoder and decoder have Convolutional, Spectral Normalization layers, and ReLU activation functions. 
    
    \begin{figure*}
\centering
\includegraphics[width=5.2in, height=2.4in]{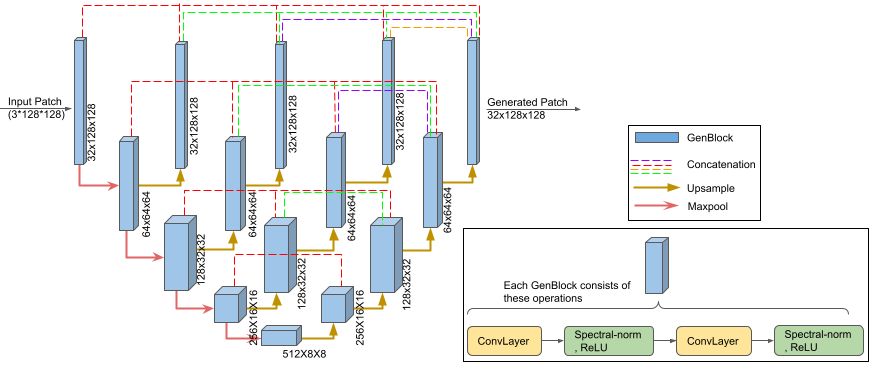} 
\vspace*{-0.4cm}
\caption{Architecture of Generator ($ex$. input patch from SIXray dataset)}
\label{fig}
\vspace*{-0.3cm}
\end{figure*}

\noindent    Let i denote the index of layers downsampling in the encoder and j index the other layers upsampling in the decoder. If we consider $x^{i,j}$ a representation of the feature maps, to be the output of a node $X^{i,j}$, then mathematically we can frame the operations on the skip-pathway as:

\noindent \textbullet \hspace{0.1cm} H:  Convolution Operation followed by an activation function. 
\\\textbullet \hspace{1pt} U:  Upsampling layer. 	\hspace{1.2cm} \textbullet \hspace{1pt} [\hspace{0pt}] $($Square  Brackets$)$:  Concatenation.
\hspace{6pt}
\begin{equation}
    x^{i,j}=
    \begin{cases}
      H(x^{i-1,j}), & \text{j=0} \\
      H\left(\left[ [x^{i,j}]^{j-1}_{k=0},  \hspace{2 pt}U(x^{i+1,j-1})\right]\right ), & \text{j $>$ 0}
    \end{cases}
  \end{equation}

\noindent \textbf{Discriminator Network(D): }
    The discriminator is used to predict/classify if the given image is real or fake. The discriminator is also used to get the
    
\begin{wrapfigure}{r}{0.5\textwidth}
  \begin{center}
    \includegraphics[width=2.7in, height=1.6in]{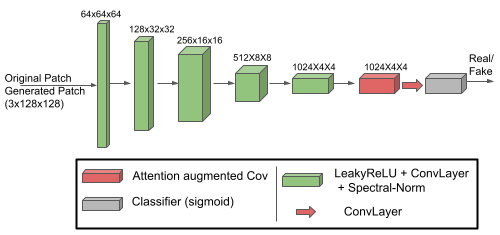} 
  \end{center}
  \caption{Architecture of Discriminator ($ex$. input patch from SIXray dataset)}
  \label{disc}
\end{wrapfigure}

       \noindent latent space representation of both the original image and reconstructed image and get inference from them. The network architecture can be seen in figure \ref{disc}. We have used multi head attention in D and spectral-normalization in both G \& D, which are one of the novel parts of this work.
       \vspace{0.2cm}

    \noindent \textbf{Attention augmented Convolution: }Convolutional operator is limited by its inability to understand global contexts \cite{hu2018squeeze, park2018bam, woo2018cbam}. However, attention mechanism can attend jointly to spatial and feature subspaces along with introducing additional feature maps \cite{zhang2019selfattention, bello2019attention}. So, we concatenate convolutional feature maps with a set of feature maps produced via self-attention, which was placed before the last convolutional layer in the discriminator. Let H represent the height, W represent the width and F represent the number of input filters of an activation map. The input tensor of shape $(H, W, F)$ is transformed into a matrix, and multi-head attention is performed over it as proposed by \cite{bello2019attention, vaswani2017attention}. For each spatial location (h,w), $N_{h}$ = 4, attention maps are computed from queries and keys as described in \cite{bello2019attention, vaswani2017attention}, which are further used to find $N_{h}$ weighted averages over the values V. The output of self attention for each head is concatenated and reshaped to match its input's spatial dimensions. The outputs are then concatenated with the output of a standard convolution operation. After experiments, an increase in performance of classification both in terms of AUC and Recall was observed when convolutions augmented with self-attention were used instead of just convolutions in discriminator. We did an ablation study for 10k random image set of SIXray dataset and we observed that the use of attention increased the AUC metric up to 0.021 and Recall metric up to 0.01.

\begin{figure}[!htb]
\begin{center}
\includegraphics[width=5.0in, height=1.8in]{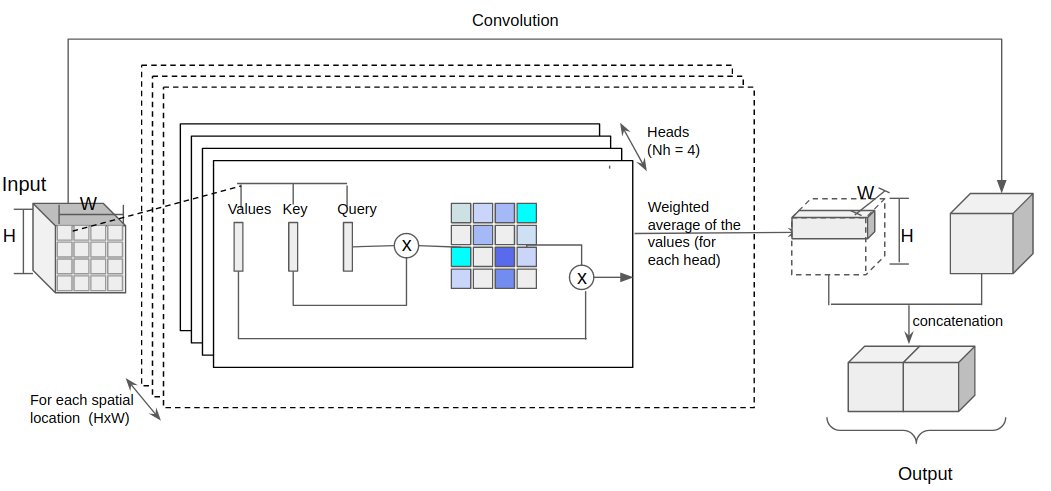}
\end{center}
\vspace*{-0.2cm}
\caption{$N_h$ attention maps are calculated for each spatial location (h,w), which then give weighted averages of values V. They are then concatenated, reshaped with same spatial dimensions as input, which is concatenated with the output of standard convolution over the input.}
\label{fig:short}
\vspace*{-0.5cm}
\end{figure}

    \noindent \textbf{Stabilizing the training of GANs:}   In training of GANs, performance control of the discriminator is a possible problem. The discriminator often inaccurately estimates the density ration in high dimensional spaces and is also unstable while training. Because of this the generator can fail to learn the multimodal structure of the target distribution. \cite{miyato2018spectral}. For a better and more stable training of our GAN, we used spectral-normalization \cite{miyato2018spectral}, in the discriminator. We also used spectral normalization in the generator, since conditioning of generator can be important for the performance of a GAN \cite{odena2018generator}. It prevents the escalation of parameter magnitudes and avoid unusual gradients, thus, avoiding problems like mode collapse\cite{zhang2019selfattention}. We found an empirically significant improvement in the training after applying spectral normalization in both generator and discriminator, at less computational costs. We also experimented with Wasserstein GANs \cite{arjovsky2017wasserstein} for this problem as an alternative, however there was no signigicant difference observed We did an ablation study for 10k random image set of SIXray dataset as mentioned in attention convolution section and the spectral normalization increased the AUC metric up to 0.008 and Recall metric up to 0.08.\\

    \noindent \textbf{Loss functions:} The two parts of GAN, namely Generator and Discriminator compete with each other eventually improving both the networks. The goal is to seek equilibrium among Generator (G) loss and Discriminator (D) loss and not minimizing the loss function. 
    Aim for G to generate as realistic image as possible preserving the contextual meaning of the image and for D to reconstruct similar latent representation as of non-anomalous images. Considering these points, we choose the following three loss functions:
   
   \noindent \emph{1. Adversarial Loss}: Inspired by adversarial loss in \cite{goodfellow2014generative}, this loss function aims to minimize the generator loss and maximize the discriminator loss \emph{min max(D,G)}. The generator will generate fake images with maximal probability of being real. The generator will ensure that the generated images are as real as possible and thus minimize the probability of image to be predicted fake by the Discriminator as in the second part of equation. The discriminator will maximize the probability of real image x and minimize the same for fake images $\hat{x}$.
$L_{adv}$ is denoted as:
\vspace*{-0.2cm}
\begin{equation}
\vspace*{-0.2cm}
    \mathcal{L}_{adv}= E_{x \sim p_{x}}[ log(D(x))] + E_{x \sim p_{x}}[ log(1 - D(\hat{x}))]
  \end{equation}
\emph{2. Contextual Loss:} Adversarial loss does not preserve the contextual meaning of the input data. Hence, to understand the input data distribution for non-anomalous images we use the Contextual loss function. For this, we apply $L1$ normalization on the real image $x$ and the generated fake image $\hat{x}$. This ensures that the
model is capable of generating images contextually similar to non-anomalous samples. The contextual loss is shown below:
\vspace*{-0.2cm}
\begin{equation}
\vspace*{-0.2cm}
    \mathcal{L}_{con}= E_{x \sim p_{x}} \vert x - \hat{x} \vert
  \end{equation}
\emph{3. Latent Loss:} 
To reconstruct the latent representation of input images $x$ very close to that of generated fake images $\hat{x}$. This will help the network produce contextually same latent representation for most of the non-anomalous samples. As
shown in Figure 3(b),the discriminator D's final convolutional layer is used to get the features of x and $\hat{x}$ and then their latent representation is reconstructed such that z = f (x) and
$\hat{z}$ = f ($\hat{x}$). Latent representation loss can be represented as:
\begin{equation}
    \mathcal{L}_{lat}= E_{x \sim p_{x}} \vert f(x) - f(\hat{x}) \vert
  \end{equation}
The total training objective then becomes a weighted sum of
the losses above:
\vspace*{-0.2cm}
\begin{equation}
\vspace*{-0.2cm}
 L = w_{1}\mathcal{L}_{adv} + w_{2}\mathcal{L}_{con} + w_{3}\mathcal{L}_{lat}
\end{equation}
where $w_{1}$, $w_{2}$ and $w_{3}$ are the weighting parameters which can be used to manage the weights for each individual loss function in the overall loss function.

    \noindent \textbf{Calculation of Anomaly scores.} For identifying anomaly in the test data we calculate a score for each image, which expresses how anomalous the given image is. This method is inspired by \cite{schlegl2017unsupervised, zenati2018efficient}, where anomaly score ${A(x)}$ for a test image $x$, is given by:
    \vspace*{-0.1cm}
    \begin{equation}
    \vspace*{-0.1cm}
   \mathcal{A}(x) = \eta\mathcal{A}_{G}(x) + (1 - \eta)\mathcal{A}_{D}(x)
   \end{equation}
    where $\mathcal{A}_{G}(x)$ represents the contextual similarity of the real input images and the images generated by the generator as mentioned above in contextual loss. 
    Therefore $\mathcal{A}_{G}(x)$ $=$  $\vert$ $x$ - $\hat{x}$ $\vert$ . Thus, while calculating anomaly score, greater the value of   $\mathcal{A}_{G}(x)$, greater is the probability of the image to have an anomaly. 
    $\mathcal{A}_{D}(x)$ in equation 6 represents the difference in latent representation of input and generated images and is measured as in latent loss section. 
    $\mathcal{A}_{D}(x)$ $=$ $\vert$ f(x) - f($\hat{x}$)$\vert$ . $\eta$ is a coefficient used to give relative weights to $\mathcal{A}_{G}(x)$ and $\mathcal{A}_{D}(x)$. 
    \noindent Let the test data set be $\mathcal{T}$, we calculate $\mathcal{A}$(x) for each $x\in\mathcal{T}$. Therefore, we get the anomaly scores vector $\mathcal{V}_\mathcal{A}$ = $\{v_i :\mathcal{A}_i(x), x_i \in T\}$. Now for normalizing, we scale these anomaly scores within the range $[0,1]$ using feature scale as explained below:
    \begin{equation}
    v'_i = \frac{v_i - min(\mathcal{V}_\mathcal{A})}{max(\mathcal{V}_\mathcal{A})-min(\mathcal{V}_\mathcal{A})}
    \end{equation}
    \noindent This $\mathcal{V'}_\mathcal{A}$ is the final anomaly score vector for test set $\mathcal{T}$ which will be used to detect anomalies.

\vspace*{-0.4cm}
\section{Experiments and Results: }
\vspace*{-0.1cm}

This section briefs on the datasets and how patches are extracted from them. Further it give details of the training objective, experimental setting and observations obtained after the experiments.
    \noindent \textbf{Datasets:} We have performed experiments on 3 datasets. From which MVTec\_AD\cite{bergmann2019mvtec} and SIXray\cite{miao2019sixray} datasets are more suitable for anomaly detection tasks, owing to their well defined anomalies. However, as CIFAR-10\cite{krizhevsky2014cifar} is a very common benchmark dataset we have produced our results for comparing with various models. The datasets are described ahead.
    
    \noindent \textbf{\emph{1. CIFAR-10}}\cite{krizhevsky2014cifar}:  A commonly used benchmark dataset  for classification. It has 60000 color images of size $32\times32$ for 10 categories. But we have used this dataset for anomaly detection by taking 1 vs all approach.
    No segmentation or patch based technique was used for pre-processing.
    \\\textbf{{\emph{2. MVTecAD }}}\cite{bergmann2019mvtec}: A publicly available dataset for bench-marking anomaly detection methods with a focus on industrial inspection. It has 5354 images divided into 15 different object and texture categories. Each category comprises a set of defect-free training images and a test set of images with various kinds of defects (with segmentation) as well as images without defects. For training and testing processes, image patches of size $256\times256$ were extracted. For images with defects, segmentation masks were used to obtain the damaged region from the complete image.
    \\\textbf{\emph{3. SIXray}}\cite{miao2019sixray}: It has \emph{1,059,231 high-resolution X-ray images} of luggage items. Less than \emph{1 percent} of images have positive labels \emph{i.e}. the luggage contains an anomalous object. There are 6 subclasses of prohibited items, namely, gun, knife, wrench, pliers, scissors and hammer. For training purpose, non-overlapping patches of size $256\times256$ were extracted  from \emph{10k, 100k, 500k} random image set. For test category, both positive labelled as well as negative labelled images were used. Publicly available annotations of around \emph{1400} images (containing an anomalous object) were used to extract the Region of Interest.\\\\\textbf{Training Objective:} Our goal in this field of anomaly detection is to maximize the classification capability of the model. Works in the past mainly focused on AUC of the ROC curve of the classification model. However, along with AUC our approach tries to increase the Recall of the model. The target domains of this work are important and sensitive fields like security, medical analysis, $etc.$ When there are wide disparities in the cost of false negatives vs. false positives, it is important to minimize one type of classification error. 
    For example: In baggage security, a safe bag classified as having a threat object can then be put through a human's investigation. However, allowing a baggage with threat  objects like pistols, knives, etc. to pass through without intervention can lead to problems. This calls for the need of better recall, which represents capability of the model to avoid false negatives. Thus, we used both AUC and Recall to evaluate our model for baggage security images (SIXray dataset).\\
    
    \noindent \textbf{Implementation Details:} The model is implemented in Python using the PyTorch framework and is trained with 32GB DDR3 RAM, Nvidia 1080 GPU and Intel 3.2GHz 4core CPU on a Linux system. For training, the image size was as mentioned for each dataset in the dataset section, the learning rate was $8e^{-3}$ and the weights of loss functions used were: w1($\mathcal{L}_{adv}$) = 1, w2($\mathcal{L}_{con}$) = 40 and w3($\mathcal{L}_{lat}$) = 1, which empirically showed the optimal performance.\\
    \\
    \textbf{Results :} For the CIFAR-10\cite{krizhevsky2014cifar} dataset, as in Table \ref{tab:my_table}, we can see that our model performs the best on all the classes. Similarly, for the MV\_Tec AD dataset\cite{bergmann2019mvtec}, we can see in table \ref{tab:my_label}, that when the results of our experiments are compared with those mentioned in DAGAN \cite{tang2020anomaly} and \cite{Yi_2020_ACCV} our model outperforms all other models in most of the classes.\\

    \begin{table}[!htb]
    \renewcommand*{\arraystretch}{0.85}

    \caption{AUC results for CIFAR-10 dataset \cite{akccay2019skip}}\label{tab:my_table}
    \vspace*{-0.2cm}
    \resizebox{\textwidth}{!}{\begin{tabular}{c c c c  c c c c c c c c c|}
     \\\hline
     Model & bird & car & cat & deer & dog & frog & horse & plane & ship & truck  \\ [0.5ex]\hline
     AnoGAN\cite{schlegl2017unsupervised} & 0.411 & 0.492 & 0.399 & 0.335 & 0.393 & 0.321 & 0.399 & 0.516 & 0.567 & 0.511 \\\hline
     EGBAD\cite{zenati2018efficient} & 0.383 & 0.514 & 0.448 & 0.374 & 0.481 & 0.353 & 0.526 & 0.577 & 0.413 & 0.555 \\\hline
      GANomaly \cite{akcay2018ganomaly} & 0.510 & 0.631 & 0.587 & 0.593 & 0.628 & 0.683 & 0.605 & 0.633 & 0.616 & 0.617 \\\hline
     \small{Skip GANomaly\cite{akccay2019skip}} 
     & 0.448 & 0.953 & 0.607 & 0.602 & 0.615 & 0.931 & 0.788 & 0.797 & 0.659 & 0.907 \\\hline
    \textbf{Proposed} & \textbf{0.982} & \textbf{0.998} & \textbf{0.981} & \textbf{0.999} & \textbf{0.989} & \textbf{1.000} & \textbf{0.992} & \textbf{1.000} & \textbf{1.000} & \textbf{0.974} \\\hline
    \end{tabular}}
    \end{table}

\vspace*{-0.4cm}
\renewcommand*{\arraystretch}{0.50}
\begin{table}[!htb]
\small
\caption{AUC on MVTec\_AD dataset \cite{tang2020anomaly} }\label{tab:my_label}
\vspace*{-0.2cm}
\hskip-1.2cm \begin{tabular}{c c c c c c c c } 
 \\\hline
 \textbf{Class} & \makecell{{AnoGAN} \\ \cite{schlegl2017unsupervised}} & \makecell{{GANomaly} \\ \cite{akcay2018ganomaly}} & \makecell{{Skip} \\ {Ganomaly}\cite{akccay2019skip}} & \makecell{{DAGAN} \\\cite{tang2020anomaly}} & \makecell{{U-Net} \\ \cite{ronneberger2015u}} & \makecell{{Patch}\\{SVDD} \\ \cite{Yi_2020_ACCV}} & \textbf{Proposed} \\ \hline
 Bottle & 0.800 & 0.794 & 0.937 & 0.983 & 0.863 & 0.986 & \textbf{0.987}
 \\\hline 
 Cable & 0.477 & 0.711 & 0.674 & 0.665 & 0.636 & 0.903 &\textbf{0.918} \\ \hline
 Capsule & 0.422 & 0.721 & 0.718 & 0.687 & 0.673 & 0.767 &\textbf{0.998} \\\hline 
 Carpet & 0.337 & 0.821 & 0.795 & 0.903 & 0.774 & 0.929 &\textbf{0.938} \\\hline 
 Grid & 0.871 & 0.743 & 0.657 & 0.867 & 0.857 & \textbf{0.946} & {0.943} \\\hline 
 HazelNut & 0.259 & 0.874 & 0.906 & \textbf{1.0} & 0.996 & 0.920 & \textbf{0.999}\\\hline 
 Leather & 0.451 & 0.808 & 0.908 & 0.844 & 0.870 & 0.909 & \textbf{0.935} \\\hline 
 Metal Nut & 0.284 & 0.694 & 0.79 & 0.815 & 0.676 & \textbf{0.940} & {0.825}  \\\hline 
  Pill & 0.711 & 0.671 & 0.758 & 0.768 & 0.781 & 0.861 & \textbf{0.91 } \\\hline
 Screw & 0.10 & 1.0 & 1.0 & 1.0 & 1.0 & 0.813 & \textbf{1.0} \\\hline
 Tile & 0.401 & 0.72 & 0.85 & 0.961 & 0.964 & 0.978 & \textbf{0.985}  \\\hline 
 Toothbrush & 0.439 & 0.700 & 0.689 & 0.950 & 0.811 & \textbf{1.0} & {0.979}  \\\hline 
 Transistor & 0.692 & 0.808 & 0.814 & 0.794 & 0.674 & \textbf{0.915} & {0.89}\\\hline 
 Wood & 0.567 & 0.920 & 0.919 & 0.979 & 0.958 & 0.965 & \textbf{0.981}\\\hline 
 Zipper & 0.715 & 0.744 & 0.663 & 0.781 & 0.750 & \textbf{0.979} & {0.890}\\\hline \hline 
 Average & 0.502 & 0.782 & 0.805 & 0.866 & 0.819 & {0.921} & \textbf{0.945}\\\hline 
 \end{tabular}
\end{table}

    \noindent Accomplishing our main goal, our model outperforms the state of the art models in baggage threat object detection. Three major training sets, were made: A, B and C, each of different size as shown in Table \ref{recall}. Hence, for  images were taken randomly in order to ensure that we cover the entire data. 
    After performing the experiments on these models for the best performance, we can see that our model outperforms both of them in terms of AUC and Recall. Its observed that our model converges in just about 20 epochs which is much faster as compared to other methods, where almost more than 35 epochs are needed. Lesser number of epochs also reduces the possibility of over-fitting. It is observed that there is a small decrease in recall as the number of images increases. We hypothesise that the decrease in Recall while still having good or increase in AUC is due to redundant data leading to model learning features or noise that is common to the data, but not unique to non-anomalous images. This variation is observed across all the methods, which further supports our hypothesis.
   
\begin{table}[!htb]
\renewcommand*{\arraystretch}{0.85}
\vspace*{-0.2cm}
\centering
\caption{ AUC and Recall on SIXray}\label{recall}
\begin{tabular}{ccccccccc}
\hline
SIXray & \multicolumn{2}{c}{GANomaly\cite{akcay2018ganomaly}}
& \multicolumn{2}{c}{{SkipGanomaly\cite{akccay2019skip}}}
& \multicolumn{2}{c}{\textbf{Proposed}}         \\
\cline{2-7}
(No. of Images)   &   AUC  &   Recall  &   AUC  &   Recall  &   AUC  &   Recall \\
\hline
A(10k)   &   0.794  &   0.51  &   0.937  &  0.68  &   \textbf{0.983}  &   \textbf{0.79}  \\
    \hline
B(100k)   &     0.800    &  0.58   &   0.954  &   0.70  &   \textbf{0.998}  &  \textbf{ 0.76 } \\
    \hline
C(500k)      &   0.998  &  0.53 &    0.998   & 0.66  &    \textbf{0.999}   &   \textbf{0.75}    \\
    \hline
\end{tabular}
\end{table}

\begin{figure}[!htb]
\centering
\begin{minipage}{.5\textwidth}
  \centering
  {\includegraphics[width=6cm]{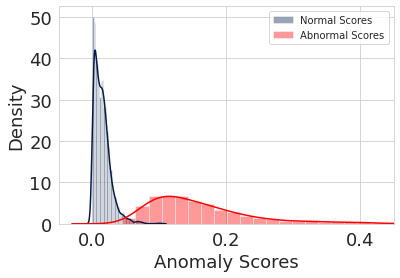} }
  \vspace*{-0.2cm}
  \caption{\small Anomaly score distribution}
  \label{fig:scr}
\end{minipage}%
\begin{minipage}{.5\textwidth}
  \centering
  {\includegraphics[width=6cm]{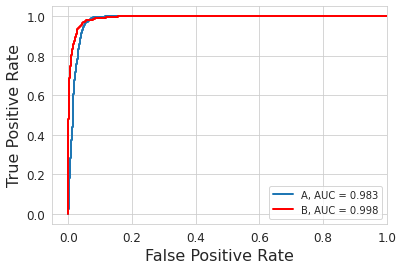} }
  \vspace*{-0.2cm}
  \caption{\small ROC Curve for A, B, C dataset of SIXray}
  \label{fig:test2}
\end{minipage}
\vspace*{-0.3cm}
\end{figure}

    \noindent Fig \ref{fig:scr} represents the anomaly scores distribution of non-anomalous test samples and anomalous test samples, which are calculated as mentioned previously. We can clearly see that the anomaly scores for anomalous images are on the higher side while that of non-anomalous samples are comparatively lower. The region, where the scores of both classes overlap, is where the network fails to identify whether the image has an anomaly or not.

    \noindent We also derived the test results of our pre-trained model with the SIXray set A, in terms of recalls, when 5 different sub-classes of threat objects in SIXray dataset were tested on separately. Region of interest of positive samples were extracted using Annotations as explained in Dataset section. We experimented with the anomalous objects in test set to check the performance of the model on different types of objects. Images of only a particular anomalous object were given and the recalls of gun, knife, pliers, scissors and wrench are 0.834, 0.931, 0.886, 0.933, 0.556 respectively. The model performs fairly well on all the categories, except wrench. This could be due to its size and physical characteristics and its similarity with non-anomalous objects.

\vspace*{-0.2cm}
\section{Conclusions}

    This paper proposes a novel method for anomaly detection. This adversarial training tackles the problem of unbalanced data. Since the training does not require the anomalies, it makes it capable to handle  variations in the threat objects. Our models examine the use of spectral normalization for stable training, the role of dense convolutional skip connections and attention for better reconstruction of image, inference learning and detecting consistency in distant features.  Our model outperforms prior work\cite{akcay2018ganomaly, akccay2019skip} on anomaly detection within X-ray secuity baggage imagery by improving the performance on SIXray dataset\cite{miao2019sixray}, by about 5\%  in AUC and about 16\% in terms of Recall of the best models previously. The proposed model achieves superior results in most of the classes in MV\_Tec AD dataset\cite{bergmann2019mvtec} compared to the state-of-the-art strategies\cite{schlegl2017unsupervised, akcay2018ganomaly, akccay2019skip, tang2020anomaly, ronneberger2015u}. This demonstrates the effectiveness of our proposal, which is, it's not limited to a particular application domain but can be beneficial for anomaly detection task in various domains. Our experiments with different dataset size with sixray also demonstrates that a very high performance can be achieved even with a very small dataset size, as compared to other methods \cite{akcay2018ganomaly}\cite{akccay2019skip} where a significant difference is seen as the dataset size is increased. This makes the model quite efficient in terms of training requirement. These sets had multiple sets in themselves.  Even within each set, different random combinations were used to make multiple sets, to avoid overfitting. The use of attention along with the improved performance also brings a higher computation power requirement, however that is manageable and can be mitigated to some extent by regularizing other parameters while still maintaining approximately the same performance. This work can be further extended by integrating it with threat object localization helping further reduce human dependency. Also, different aspects such as continual learning could be explored to help the model learn to adapt, to the changes in bags and objects.\\

\noindent \textbf{Conflict of interest statement:}
We declare that there is no conflict of interest with any person or organization professional or personal that could influence our work.

\bibliography{sn-bibliography}


\end{document}